\documentclass{article}

\usepackage[final,nonatbib]{nips_2017}
\usepackage{graphicx}

\usepackage{amssymb,amsmath}

\usepackage[utf8]{inputenc} \usepackage[T1]{fontenc}    \usepackage{hyperref}       \usepackage{url}            \usepackage{booktabs}       \usepackage{amsfonts}       \usepackage{nicefrac}       \usepackage{microtype}      \usepackage{color}  
\usepackage{wrapfig}

\title{Learning Wasserstein Embeddings}

\author{  Nicolas Courty\thanks{All three authors contributed equally.} \\
 Université de Bretagne Sud,\\
  IRISA, UMR 6074, CNRS,\\
  \texttt{courty@univ-ubs.fr} \\
  \And
    Rémi Flamary$^*$\\
 Université Côte d’Azur, OCA\\
  Lagrange, UMR 7293, CNRS\\
  \texttt{remi.flamary@unice.fr} \\
  \And
      Mélanie Ducoffe$^*$\\
 Université Côte d’Azur,\\
 I3S, UMR 7271, CNRS\\
  \texttt{melanie.ducoffe@unice.fr} \\
                                          }

\begin{document}

\maketitle

\begin{abstract}
 The Wasserstein distance received a lot of attention recently in the community of machine learning, especially 
 for its principled way of comparing distributions. It has found numerous applications in several hard problems, such as domain 
 adaptation, dimensionality reduction or generative models. However, its use is still limited by a heavy computational cost. 
 Our goal is to alleviate this problem by providing an approximation mechanism
 that allows to break its inherent complexity. 
 It relies on the search of an embedding where the Euclidean distance mimics the Wasserstein distance. We show that such an 
 embedding can be found with a siamese architecture associated with a decoder network that allows to move from the 
 embedding space back to the original input space. Once this embedding has been found, computing optimization problems
 in the Wasserstein space ({\em e.g.} barycenters, principal directions or even archetypes) can be conducted extremely fast. Numerical experiments 
 supporting this idea are conducted on image datasets, and show the wide potential benefits of our method. 
\end{abstract}

\section{Introduction}
\label{sec:intro}

The Wasserstein distance  is a powerful tool based on the theory of optimal transport to compare data
distributions with wide applications in image processing, computer vision and machine learning~\cite{Kolouri17}. 
In a context of machine learning, it has recently found numerous applications, {\em e.g.} domain adaptation~\cite{courty2017}, 
 word embedding~\cite{huang2016} or generative models~\cite{arjovsky17a}. Its power comes from two major reasons: {\em i)} it 
 allows to operate on empirical data distributions in a non-parametric way {\em ii)} the geometry of the underlying space can be 
 leveraged to compare the distributions in a geometrically sound way. The space of probability measures equipped with the Wasserstein 
 distance can be used to construct objects of interest such as
 barycenters~\cite{agueh2011barycenters} or geodesics~\cite{seguy2015principal} that 
 can be used in data analysis and mining tasks.

More formally, let $X$ be a metric space endowed with a metric $d_X$. Let $p\in(0,\infty)$ and $\mathcal{P}_p(X)$ the space of all Borel
probability measures $\mu$ on $X$ with finite moments of order p, {\em i.e.} $\int_X d_X(x,x_0)^pd\mu(x) < \infty$ for all $x_0$ in $X$.
The p-Wasserstein distance between $\mu$ and $\nu$ is defined as:
 \begin{equation}
 W_p(\mu,\nu) = \left(\inf_{\pi \in \Pi(\mu,\nu)} \iint_{X\times X} d(x,y)^pd\pi(x,y)\right)^{\frac{1}{p}}.
 \label{eq:Wasserstein}
 \end{equation}
Here, $\Pi(\mu,\nu)$ is the set of probabilistic couplings $\pi$ on $(\mu,\nu)$. As such, for every Borel subsets $A \subseteq X$, we have
that $\mu(A)=\pi(X\times A)$ and $\nu(A)=\pi(A\times X)$. It is well known that $W_p$ defines a metric over $\mathcal{P}_p(X)$ as long
as $p\geq1$ (e.g.~\cite{Villani09}, Definition 6.2).

When $p=1$, $W_1$ is also known as Earth Mover's distance (EMD) or Monge-Kantorovich distance. The geometry of ($\mathcal{P}_p(X)$, $W_1(X)$)
has been thoroughly studied, and there exists several works on computing EMD for point sets in $\mathbb{R}^k$ (e.g.~\cite{Shirdhonkar08}). However, in a number of
applications the use of $W_2$ (a.k.a {\em root mean square bipartite matching distance}) is a more natural distance arising in computer vision~\cite{Bonneel2015}, computer graphics~\cite{Bonneel11,deGoes2012, Solomon2015,Bonneel2016} or machine learning~\cite{Cuturi14,courty2017}. See~\cite{deGoes2012} for a discussion on
the quality comparison between $W_1$ and $W_2$.

Yet, the deployment of Wasserstein distances in a wide class of applications is somehow limited, especially because of an heavy computational burden.
In the discrete version of the above optimisation problem, the number of variables scale quadratically with the number of samples in the distributions, and 
solving the associated linear program with network flow algorithms  is known to
have a cubical complexity. While recent strategies implying slicing
technique~\cite{Bonneel2015,Kolouri16a}, entropic
regularization~\cite{Cuturi13,benamou2015iterative,solomon2015convolutional} or involving stochastic optimization~\cite{genevay2016}, have emerged, the cost of
computing pairwise Wasserstein distances between a large number of distributions (like an image collection) is prohibitive. This is all the more true if one 
considers the problem of computing barycenters~\cite{Cuturi14,benamou2015iterative} or population means. A recent attempt by Staib and colleagues~\cite{Staib17} 
use distributed computing for solving this problem in a scalable way.

We propose in this work to learn an Euclidean embedding of distributions where the Euclidean norm approximates the  Wasserstein distances. Finding such an embedding 
enables the use of standard Euclidean methods in the embedded space and significant speedup in pairwise Wasserstein distance computation, or construction of objects of interests such 
as barycenters.  The embedding is expressed as a deep neural network, and is learnt with a strategy similar to those of Siamese networks~\cite{chopra2005learning}. We also show that 
simultaneously learning the inverse of the embedding function is possible and allows for a reconstruction of a probability distribution from the embedding.
We first start by describing existing works on Wasserstein space embedding. We then proceed by presenting our learning framework and give proof of concepts and empirical results on existing datasets.

\section{Related work}
\label{sec:rel}

\paragraph{Metric embedding} The question of metric embedding usually arises in the context of approximation algorithms.
Generally speaking, one seeks a new representation (embedding) of data at hand in a new space where the distances from
the original space are preserved. This new representation should, as a positive side effect, offers computational ease for
time-consuming task (e.g. searching for a nearest neighbor), or interpretation facilities (e.g. visualization of high-dimensional datasets).
More formally, given two metrics spaces $(X,d_X)$ and $(Y,d_y)$ and $D \in [1,\infty)$, a mapping $\phi : X \rightarrow Y$ is an
embedding with distortion at most $D$ if there exists a coefficient $\alpha \in (0,\infty)$ such that $\alpha d_X(x,y) \leq d_Y(\phi(x),\phi(y)) \leq D\alpha d_X(x,y)$.
Here, the $\alpha$ parameter is to be understood as a global scaling coefficient. The {\bf distortion} of the mapping is the infimum over all
possible $D$ such that the previous relation holds. Obviously, the lower the $D$, the better the quality of the embedding is. It should be noted
that the existence of exact ({\bf isometric}) embedding ($D=1$) is not always guaranteed but sometimes possible. Finally, the embeddability of a
metric space into another is possible if there exists a mapping with constant distortion. A good introduction on metric embedding can be found
in~\cite{matouvsek2013lecture}.

\paragraph{Theoretical results on Wasserstein space embedding}
Embedding Wasserstein space in normed metric space is still a theoretical and open questions~\cite{Matousek11}. Most of the theoretical
guarantees were obtained with $W_1$. In the simple case where $X=\mathbb{R}$, there exists an isometric embedding with $L_1$ between two absolutely 
continuous ({\em wrt.} the Lebesgue measure) probability measures $\mu$ and $\nu$ given by their by their cumulative distribution functions $F_{\mu}$ and $F_{\nu}$, {\em i.e.}
$W_1(\mu,\nu)=\int_\mathbb{R} |  F_{\mu}(x) -F_{\nu}(x) | dx$. This fact has been exploited in the computation of sliced Wasserstein distance~\cite{Bonneel2015,Kolouri2016}.
Conversely, there is no known isometric embedding for pointsets in $[n]^k=\{1,2,\hdots,n\}^k$, {\em i.e.} regularly sampled grids in $\mathbb{R}^k$, but best known 
distortions are between $O(k \log n)$ and $\Omega (k+\sqrt{\log n})$~\cite{Charikar02,Indyk03,Khot2006}. Regarding $W_2$, recent results~\cite{andoni16}
have shown there does not exist meaningful embedding over $\mathbb{R}^3$ with constant approximation. Their results show notably that an embedding of
pointsets of size $n$ into $L_1$ must incur a distortion of $O(\sqrt{\log n})$. Regarding our choice of $W_2^2$, there does not exist embeddability results up to 
our knowledge, but we show that, for a population of locally concentrated measures, a good approximation can be obtained with our technique.  We now turn to existing 
methods that consider local linear approximations of the transport problem.

\paragraph{Linearization of Wasserstein space}
Another line of work~\cite{Wang2013,Kolouri16} also considers the Riemannian structure of the Wasserstein space to provide meaningful linearization by projecting onto
the tangent space. By doing so, they notably allows for faster computation of
pairwise Wasserstein distances (only $N$ transport computations instead of
$N(N-1)/2$ with $N$ the number of samples in the dataset) and allow for
statistical analysis of the embedded data. They proceed by specifying a template
element and compute, from particle approximations of the data, linear transport
plans with this template element, that  allow to derive an embedding used for
analysis. Seguy and Cuturi~\cite{seguy2015principal} also proposed a similar pipeline, based on velocity field, but without relying on an implicit embedding. It is to be noted that for data in 2D, such as images, the use of cumulative Radon transform also allows for an embedding which can be used for interpolation or analysis~\cite{Bonneel2015,Kolouri16a}, by exploiting the exact solution of the optimal transport in 1D through cumulative distribution functions.

Our work is the first to propose to learn a generic embedding rather than constructing it from explicit approximations/transformations of the data and analytical operators such as Riemannian Logarithm maps. As such, our formulation is generic and adapts to any type of data. Finally, since the mapping to the embedded space is constructed explicitly, handling unseen data does not require to compute new optimal transport plans or optimization, yielding extremely fast computation performances, with similar approximation performances.

\section{Deep Wasserstein Embedding (DWE)}
\label{sec:method}

\subsection{Wasserstein learning and reconstruction with siamese networks}
\label{sec:wasslearning}
We discuss here how our method, coined DWE for Deep Wasserstein Embedding, learns in a supervised way a new representation
of the data. To this end we need a pre-computed dataset that consists of pairs of
histograms $\{x^1_i,x^2_i\}_{i\in 1,...,n}$ of dimensionality $d$ and
their corresponding $W_2^2$ Wasserstein distance $\{y_i=W_2^2(x^1_i,x^2_i)\}_{i\in 1,...,n}$. 
One immediate way to solve the problem would be to concatenate the
samples $x^1$ and $x^2$ and learn a deep network that predicts
$y$. This would work in theory but it would prevent us from interpreting the
Wasserstein space and it is not by default symmetric which is a key
property of the Wasserstein distance. 

Another way to encode this symmetry and to have a meaningful embedding that can be used more
broadly is to use a Siamese neural network \cite{bromley1994signature}. Originally designed 
for metric learning purpose and similarity learning (based on labels), this type of architecture is usually defined by 
replicating a network  which takes as input two samples from the same learning set, and learns a mapping to new space 
with a contrastive loss. It has mainly been used in computer vision, with successful  applications to face recognition~\cite{chopra2005learning}
 or one-shot learning for example~\cite{koch2015siamese}. Though its capacity to learn meaningful embeddings has been highlighted in~\cite{weston2012deep}, 
 it has never been used, to the best of our knowledge, for mimicking a specific distance that exhibits computation challenges. 
This is precisely our objective here.

\begin{figure}[t]
 \centering
 \includegraphics[width=1\textwidth]{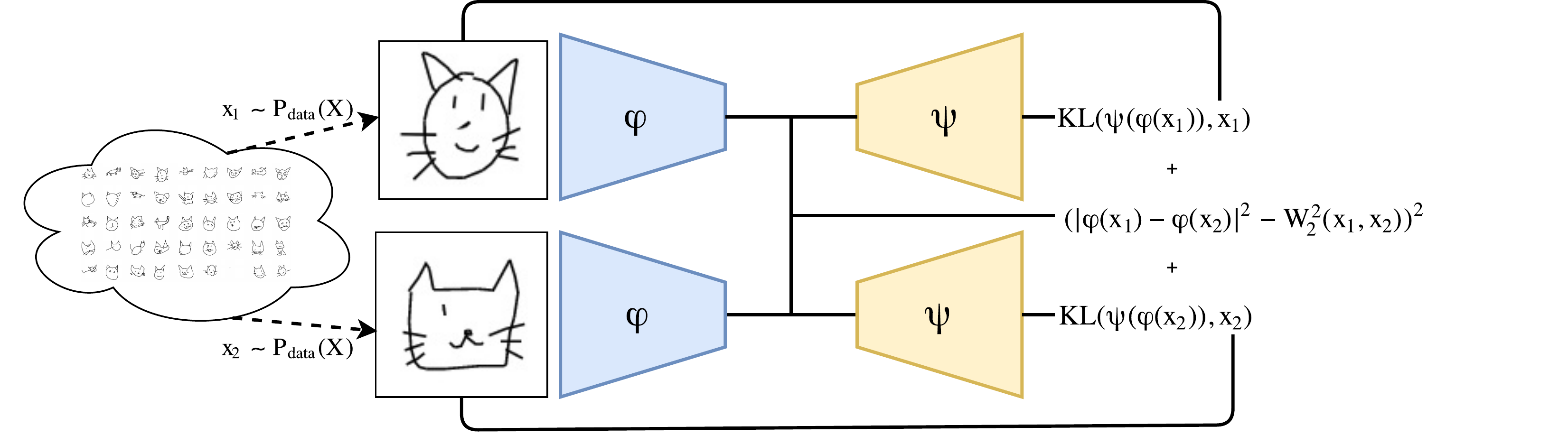}
 \caption{Architecture of the Wasserstein Deep Learning: two samples are drawn from the data distribution and set as input of the same network ($\phi$) that computes the embedding. The embedding is learnt such that the squared Euclidean distance in the embedding mimics the Wasserstein distance. The embedded representation of the data is then decoded with a different network ($\psi$), trained with a Kullback-Leibler divergence loss.}
 \label{fig:DWE}
\end{figure}

We propose to learn and embedding network $\phi$ that takes as input a histogram and
project it in a given Euclidean space of $\mathbb{R}^p$. In practice, this embedding should  mirror the geometrical property of the
Wasserstein space. We also propose to regularize the computation this of this embedding by adding a reconstruction loss based on 
a decoding network $\psi$. This has two important impacts: First we
observed empirically that it eases the learning of the embedding and improves
the generalization performance of the network (see experimental results) by forcing the embedded representation to catch sufficient information 
of the input data to allow a good reconstruction. This type of autoencoder regularization loss has been discussed in~\cite{yu2013embedding}
in the different context of embedding learning. Second, disposing of.a decoder network allows the interpretation of the results, which is 
of prime importance in several data-mining tasks (discussed in the next subsection).

An overall picture depicting the whole process is given in Figure~\ref{fig:DWE}. The global objective function reads 
\begin{equation}
  \label{eq:wass_learning}
  \min_{\phi,\psi}\quad \sum_i \left\|
    \|\phi(x^1_i)-\phi(x^2_i)\|^2-y_i\right\|^2+\lambda \sum_i\text{KL}(\psi(\phi(x_i^1)),x_i^1)+\text{KL}(\psi(\phi(x_i^2)),x_i^2)
\end{equation}
where $\lambda>0$ weights the two data fitting terms and $\text{KL}(,)$ is the
Kullbach-Leibler divergence. This choice is motivated by the fact that the Wasserstein metric operates on probability distributions.

\subsection{Wasserstein data mining in the embedded space}
\label{sec:datamining}

Once the functions $\phi$ and $\psi$ have been learned, several data mining tasks can be operated in the Wasserstein space. 
We discuss here the potential applications of our computational scheme  and its
wide range of applications on problems where 
the Wasserstein distance plays an important role. Though our method is not an exact Wasserstein estimator, we empirically show 
in the numerical experiments that it performs very well and competes favorably with other classical computation strategies.

\paragraph{Wasserstein barycenters~\cite{agueh2011barycenters,Cuturi14,Bonneel2016}.} Barycenters in Wasserstein space were first discussed by Agueh and Carlier~\cite{agueh2011barycenters}. Designed through an analogy with barycenters in a Euclidean space, the Wasserstein barycenters of a family of measures are defined as minimizers of 
  a weighted sum of squared Wasserstein distances. In our framework, barycenters can be obtained as
\begin{equation}
  \label{eq:wassbary}
  \bar x =\arg\min_x \sum_i \alpha_i W(x,x_i)\approx \psi(\sum_i \alpha_i\phi(x_i)),
\end{equation}
where $x_i$ are the data samples and the weights $\alpha_i$ obeys the following constraints: $\sum_i \alpha_i=1$ and
$\alpha_i>0$. Note that when we have only two samples, the barycenter
corresponds to a Wasserstein interpolation between the two distributions with
$\boldsymbol{\alpha}=[1-t,t]$ and $0\leq t \leq 1$ \cite{santambrogio2014introduction}. When the weights are uniform and the whole data collection is considered, the barycenter
is the Wasserstein population mean, also known as Fr{\'e}chet mean~\cite{bigot2017geodesic}. 

\paragraph{Principal Geodesic Analysis in Wasserstein space~\cite{seguy2015principal,bigot2017geodesic}.} 
 PGA, or Principal Geodesic Analysis, has first been introduced by Fletcher et al.~\cite{Fletcher04}. It can be seen as a generalization of PCA on general Riemannian manifolds. Its goal is to find a set of directions, called geodesic directions or principal geodesics,  that best encode the statistical variability of the data. It is possible to define PGA by making an analogy with PCA. Let $x_i \in \mathbb{R}^n$ be a set of elements, the classical PCA amounts to {\em i)} find $\overline x$ the mean of the data and subtract it to all the samples {\em ii)} build recursively a subspace $V_k = \mathtt{span}(v_1,\cdots,v_k)$ by solving the following maximization problem: 
\begin{equation}
	v_1  = \text{argmax}_{|v|=1} \sum^n_{i=1}(v.x_i)^2,\;\;\;v_k  = \text{argmax}_{|v|=1} \sum^n_{i=1}\left((v.x_i)^2+\sum^{k-1}_{j=1}(v_j.x_i)^2\right).
\end{equation}
Fletcher gives a generalization of this problem for complete geodesic spaces by extending three important concepts: {\bf variance} as the expected value of the squared Riemannian distance from mean, {\bf Geodesic subspaces} as a portion of the manifold generated by principal directions, and a {\bf projection} operator onto that geodesic submanifold. The space of probability distribution equipped with the Wasserstein metric ($\mathcal{P}_p(X)$, $W^2_2(X)$) defines a geodesic space with a Riemannian structure~\cite{santambrogio2014introduction}, and an application of PGA is then an appealing tool for analyzing distributional data. However, as noted in~\cite{seguy2015principal,bigot2017geodesic}, a direct application of Fletcher's original algorithm is intractable because $\mathcal{P}_p(X)$ is infinite dimensional and there is no analytical expression for the exponential or logarithmic maps allowing to travel to and from  the corresponding Wasserstein tangent space. We propose a novel PGA approximation as the following procedure: {\em i)} find $\overline x$ the approximate Fr{\'e}chet mean of the data as $\overline x=\frac{1}{N} \sum_i^N \phi(x_i)$ and subtract it to all the samples {\em ii)} build recursively a subspace $V_k = \mathtt{span}(v_1,\cdots,v_k)$ in the embedding space ($v_i$ being of the dimension of the embedded space) by solving the following maximization problem: 
\begin{equation}
	v_1  = \text{argmax}_{|v|=1} \sum^n_{i=1}(v.\phi(x_i))^2,\;\;\;v_k  = \text{argmax}_{|v|=1} \sum^n_{i=1}\left((v.\phi(x_i))^2+\sum^{k-1}_{j=1}(v_j.\phi(x_i))^2\right).
\end{equation}
which is strictly equivalent to perform PCA in the embedded space. Any reconstruction from the corresponding subspace to the original space is conducted through $\psi$. 
We postpone a detailed analytical study of this approximation to subsequent works, as it is beyond the goals of this paper.

\paragraph{Other possible methods.} As a matter of facts, several other methods that operate on distributions can benefit from our approximation scheme. 
Most of those methods are the transposition of their Euclidian counterparts
in the embedding space.  Among them, clustering methods, such as Wasserstein k-means~\cite{Cuturi14}, are readily adaptable to our framework. 
Recent works have also highlighted the success of using Wasserstein distance in dictionary learning~\cite{rolet2016fast} or archetypal Analysis~\cite{wu2017statistical}.

\section{Numerical experiments}
\label{sec:numexpe}

{
In this section we evaluate the performances of our method on grayscale images normalized as histograms. Images are offering a nice testbed 
because of their dimensionality and because large datasets are frequently available in computer vision.}

\subsection{Architecture for DWE between grayscale images}
The framework of our approach as shown in Fig \ref{fig:DWE} consists of an encoder $\phi$ and a decoder $\psi$
composed as a cascade. The encoder produces the representation of input images $h=\phi(x)$.
The architecture used for the embedding $\phi$ consists in 2
convolutional layers with  ReLu activations: first a convolutional layer of 20 filters with a kernel of size 3 by 3, then a convolutional layer
of 5 filters of size 5 by 5. The convolutional layers are followed by two linear dense layers respectively of size 100 and the final layer of size $p=50$. 
The architecture for the reconstruction
$\psi$ consists in a dense layer of output 100 with ReLu activation, followed by a dense layer of output 5*784. We reshape the layer to 
map the input of a convolutional layer: we reshape the output vector into a (5,28,28) 3D-tensor. Eventually, we invert the convolutional layers of $\phi$
with two convolutional layers: first a convolutional layer of 20 filters with ReLu activation and a kernel of size 5 by 5, followed by a second layer with 1 filter, with a kernel of size
3 by 3. Eventually the decoder outputs a reconstruction image of shape 28 by 28.
In this work, we only consider grayscale images, that are normalized to represent probability distributions.
Hence each image is depicted as an histogram. In order to normalize the decoder reconstruction
we use a softmax activation for the last layer. 

All the dataset considered are handwritten data and hence holds an inherent sparsity.
In our case, we cannot promote the output sparsity through a convex L1
regularization because the softmax outputs positive values only
and forces the sum of the output to be 1.  Instead, we apply a $\ell_p^p$ pseudo
-norm regularization with $p=1/2$ on the reconstructed image, {which promotes sparse output 
and allows for a sharper reconstruction of the images} \cite{gasso2009recovering}.
   
\subsection{MNIST digit dataset}
\label{sec:mnist}

\paragraph{Dataset and training.}
Our first numerical experiment is performed on the well known MNIST digits
dataset. This dataset contains $28\times28$ images from 10 digit classes 
In order to
create the training dataset we draw randomly one million pairs of indexes from the 60 000
training samples  and compute the exact Wasserstein distance for
quadratic ground metric using the POT toolbox \cite{flamary2017pot}. All those
pairwise distances can be computed in an embarrassingly parallel scheme
(1h30 on 1 CPU).
Among this million, 700 000 are used for learning the neural network, 200 000
are used for validation and 100 000 pairs are used for testing purposes. The DWE model is learnt on a standard NVIDIA GPU node and takes around 1h20 with a
stopping criterion computed from on a validation set.

\begin{figure}[t]
  \centering
  $\begin{array}{cc}
  \raisebox{-.5\height}{\includegraphics[width=.4\linewidth]{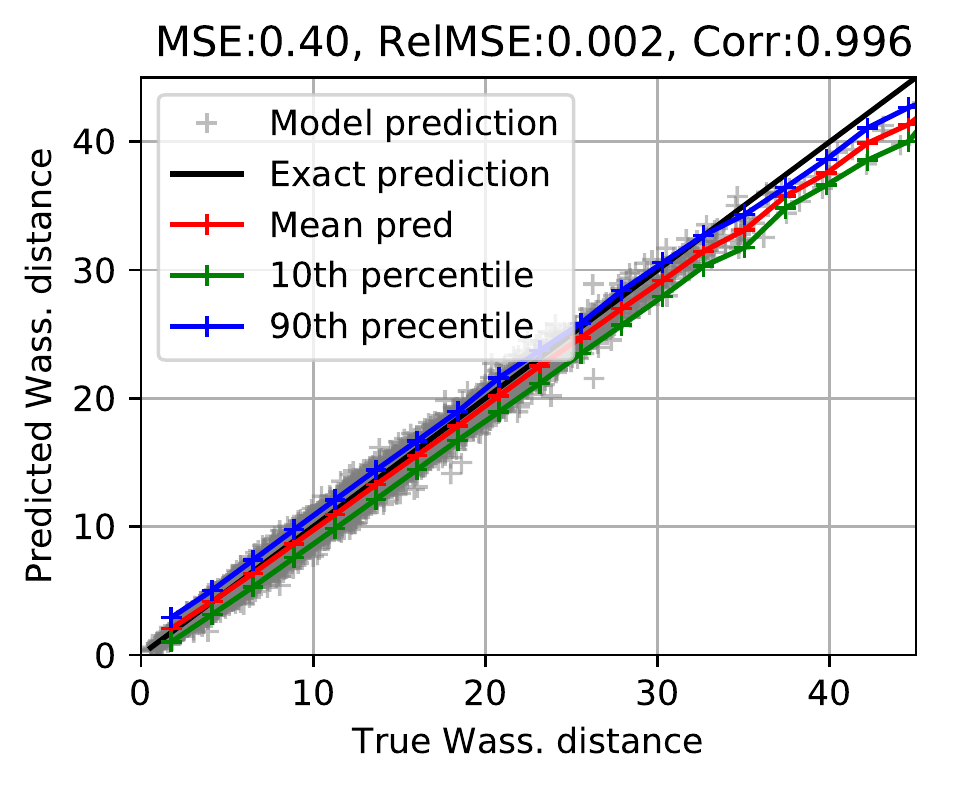}} &
  \begin{tabular}{|l|r|}\hline
   Method               &  $W^2_2$/sec \\ \hline \hline
 LP network flow (1 CPU)          &  192 \\ \hline
       DWE Indep. (1 CPU)       &  3 633 \\ 
    DWE Pairwise  (1 CPU)   &  213 384 \\\hline 
    DWE Indep. (GPU)      &  233 981 \\ 
    DWE Pairwise (GPU)     &  10 477 901 \\ \hline 
    \end{tabular}
    \end{array}$
   \caption{Prediction performance on the MNIST dataset. (Figure) The test performance are as
  follows: MSE=0.40, Relative MSE=0.002 and Correlation=0.996. (Table)
  Computational performance of $W^2_2$ and DWE given as average number of $W^2_2$
  computation per seconds for different configurations.}
  \label{fig:pred_mnist}
\end{figure}

\paragraph{Numerical precision and computational performance} The true and predicted values for the
Wasserstein distances are given in Fig. \ref{fig:pred_mnist}. We can
see that we reach a good precision with a test MSE of 0.4 and a
relative MSE of 2e-3. The correlation is of 0.996 and the quantiles
show that we have a very small uncertainty with only a slight bias for large
values where only a small
number of samples is available. This results show that a good approximation of
the $W^2_2$ can be performed by our approach ($\approx$1e-3 relative error).

Now we investigate the ability of our
approach to compute $W^2_2$ efficiently. To this end we compute the average speed of
Wasserstein distance computation on test dataset to estimate the number of $W^2_2$ computations per
second in the Table of Fig.~\ref{fig:pred_mnist}. Note that there are 2 ways to compute the
$W^2_2$ with our approach denoted as Indep and Pairwise. This comes from the
fact that our $W^2_2$ computation is basically a squared Euclidean norm in the
embedding space. The first computation measures the time to compute the $W^2_2$
between independent samples by projecting both in the embedding and computing
their distance. The second computation aims at computing all the pairwise $W^2_2$
between two sets of samples and this time one only needs to project the samples
once and compute all the pairwise distances, making it more efficient. Note that
the second approach would be the one used in a retrieval problem where one would
just embed the query and then compute the distance to all or a selection of
the dataset to find a Wasserstein nearest neighbor for instance. The speedup achieved by our method is
very impressive even on CPU with speedup of x18 and x1000 respectively for Indep
and Pairwise. But the GPU allows an even larger speedup of
respectively x1000 and x500 000 with respect to a state-of-the-art  C
compiled Network Flow LP solver of the POT Toolbox \cite{flamary2017pot,Bonneel11}. 
Of course this speed-up comes at the price of a time-consuming learning phase, 
which makes our method better suited for mining large scale datasets
and online applications.

\begin{figure}[t]
  \centering
  \includegraphics[width=\linewidth]{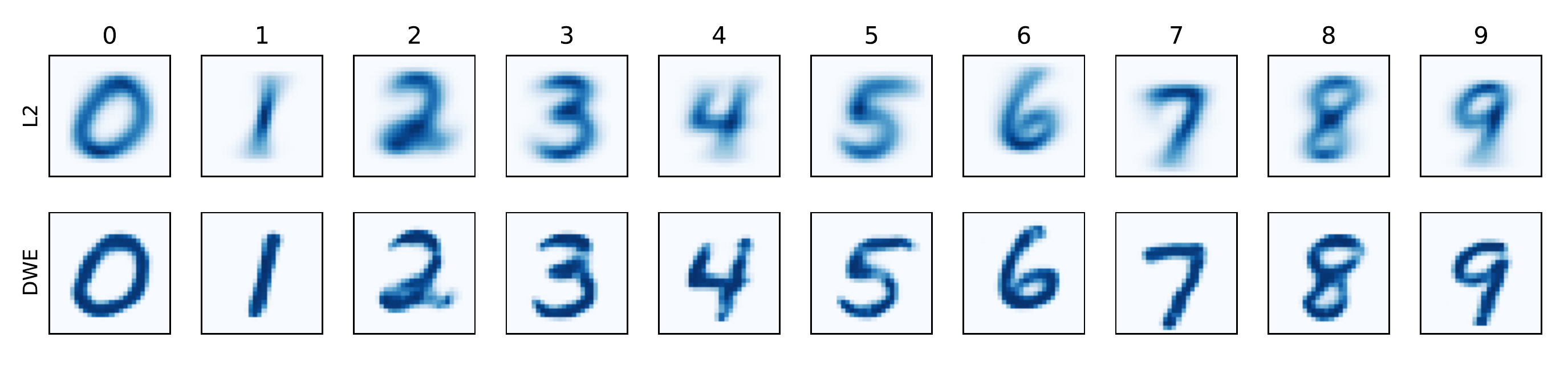}
  \caption{Barycenter estimation on each class of the MNIST dataset
    for squared Euclidean distance (L2) and Wasserstein Deep Learning (DWE).}
  \label{fig:bary}
\end{figure}

\paragraph{Wasserstein Barycenters}
Next we evaluate our embedding on the task of computing Wasserstein Barycenters for each class of the MNIST
dataset. We take 1000 samples per class from the test dataset and compute their uniform weight Wasserstein Barycenter using
Eq. \ref{eq:wassbary}. The resulting barycenters and their Euclidean means are
reported in Fig.~\ref{fig:bary}. Note that not only those barycenters are sensible but also
conserve most of their sharpness which is a problem that  occurs for regularized barycenters \cite{solomon2015convolutional,benamou2015iterative}.
The computation of those barycenters is also very efficient since it requires only 20ms per barycenter (for 1000 samples) and 
its complexity scales linearly with the number of samples.

\begin{figure}[t]
  \def \barywidth{.16}
  \setlength{\tabcolsep}{1pt}
  \centering
\begin{tabular}[t]{|c|c||c|c||c|c|}\hline
\multicolumn{2}{|c||}{Class 0} &\multicolumn{2}{|c||}{Class 1}
&\multicolumn{2}{|c|}{Class 4}\\ 
L2 & DWE & L2& DWE & L2 & DWE\\\hline
  \includegraphics[width=\barywidth\linewidth]{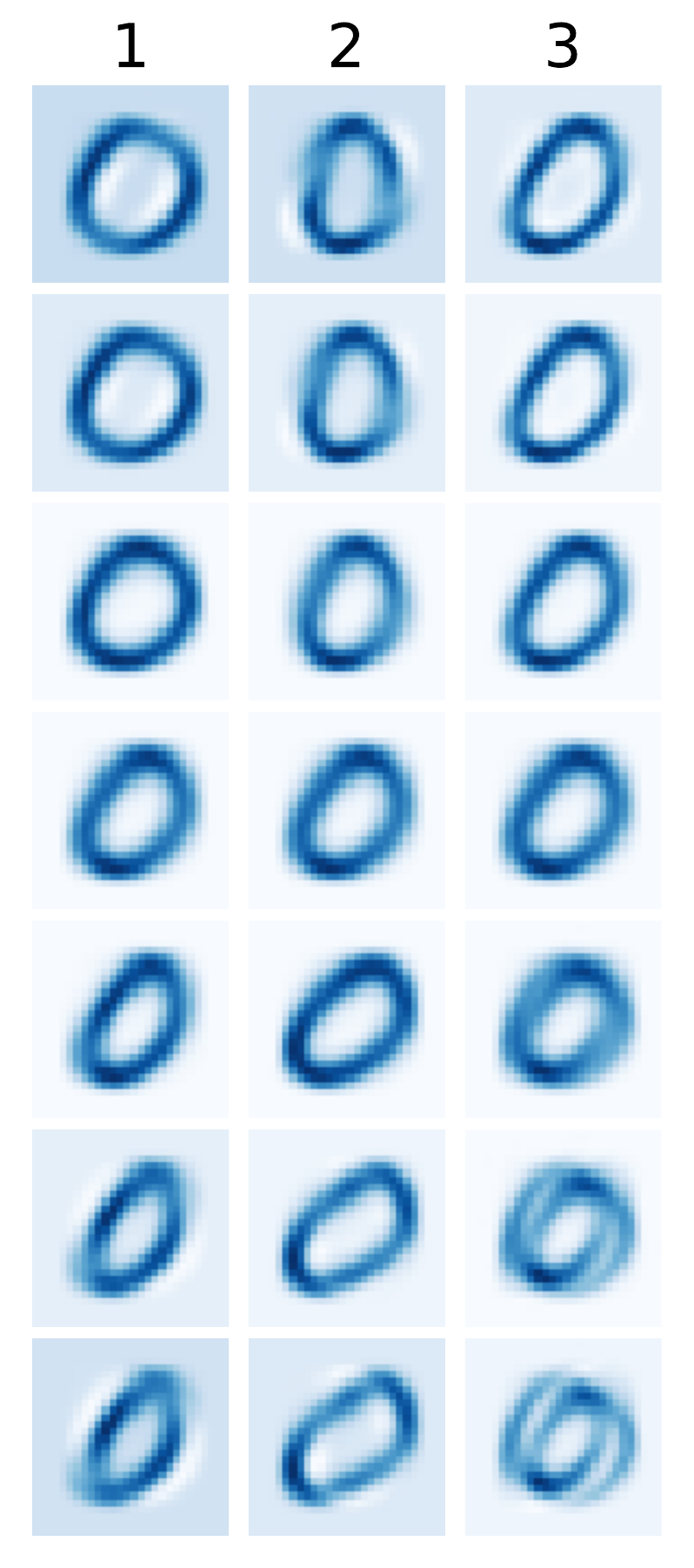}&
  \includegraphics[width=\barywidth\linewidth]{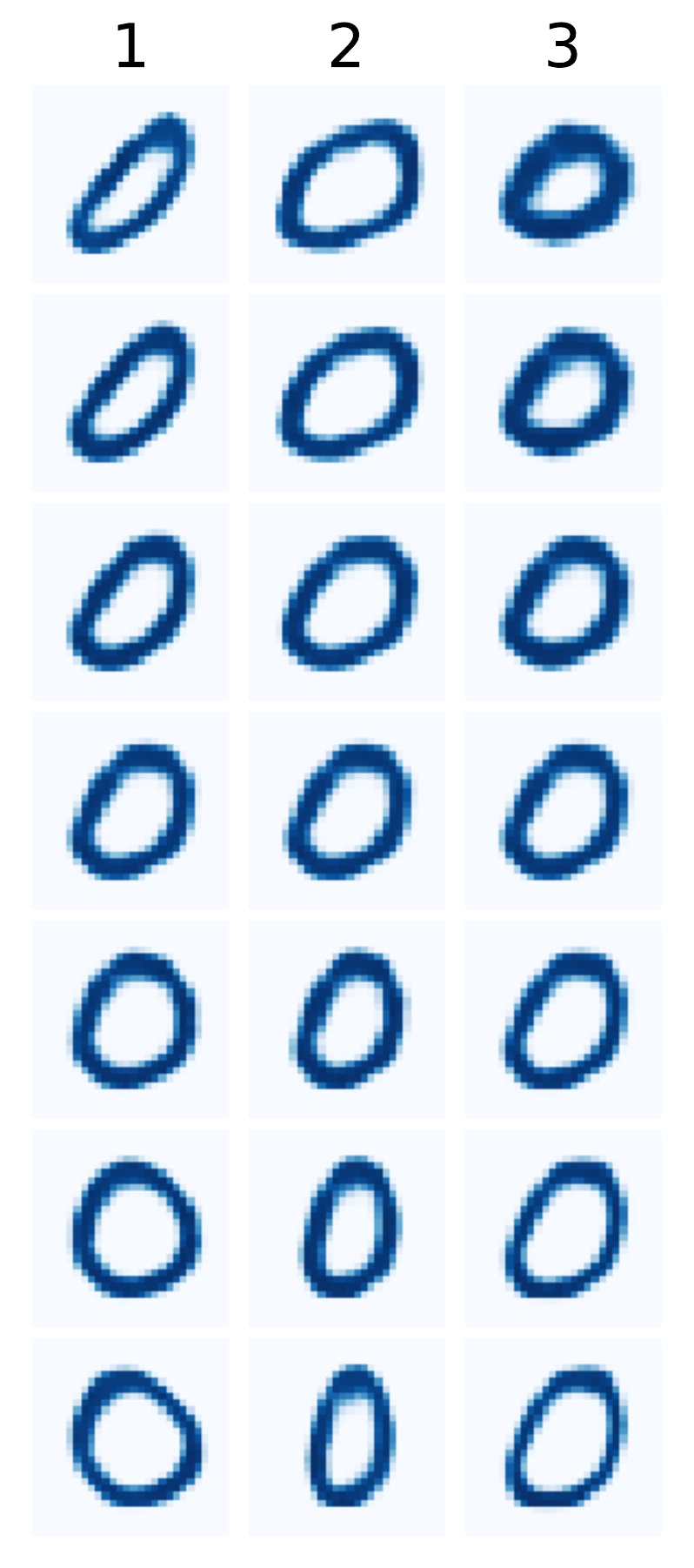}&
  \includegraphics[width=\barywidth\linewidth]{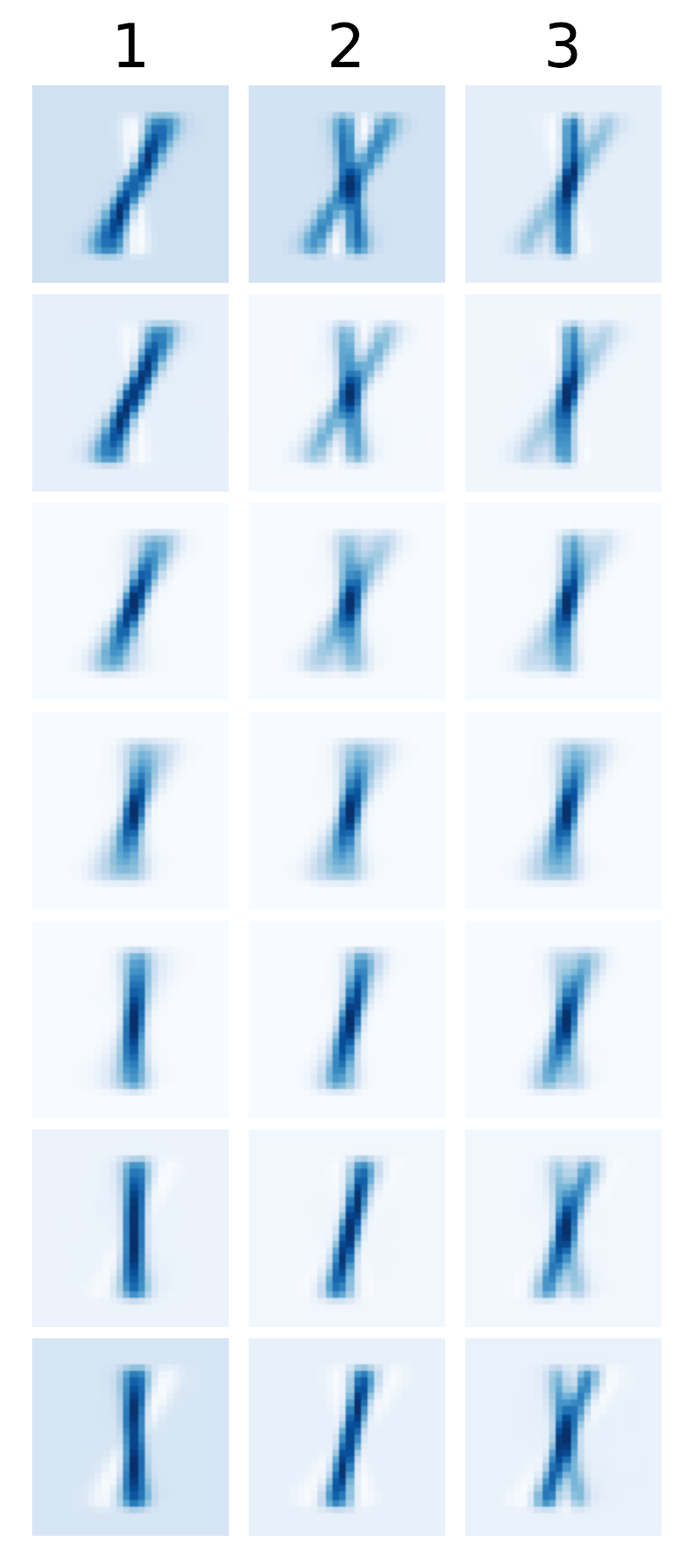}&
  \includegraphics[width=\barywidth\linewidth]{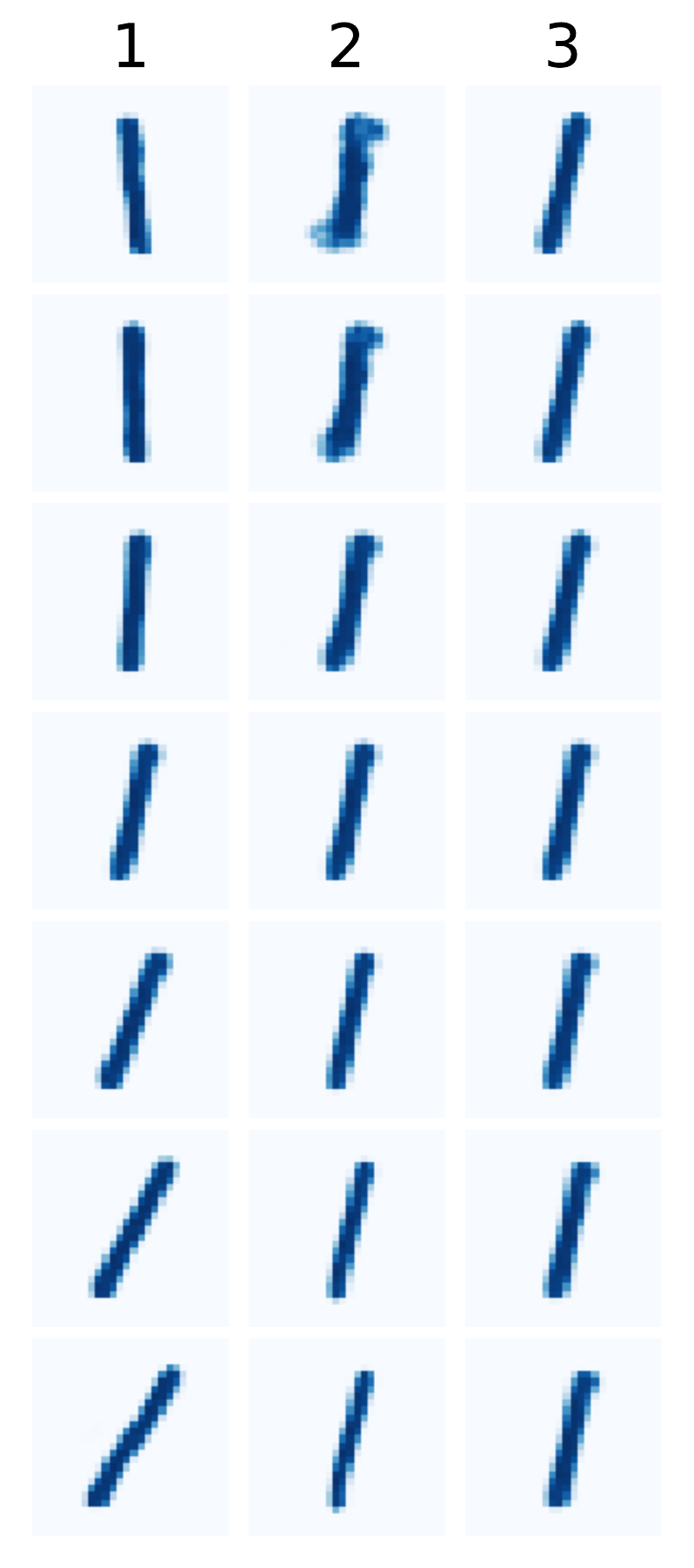}&
  \includegraphics[width=\barywidth\linewidth]{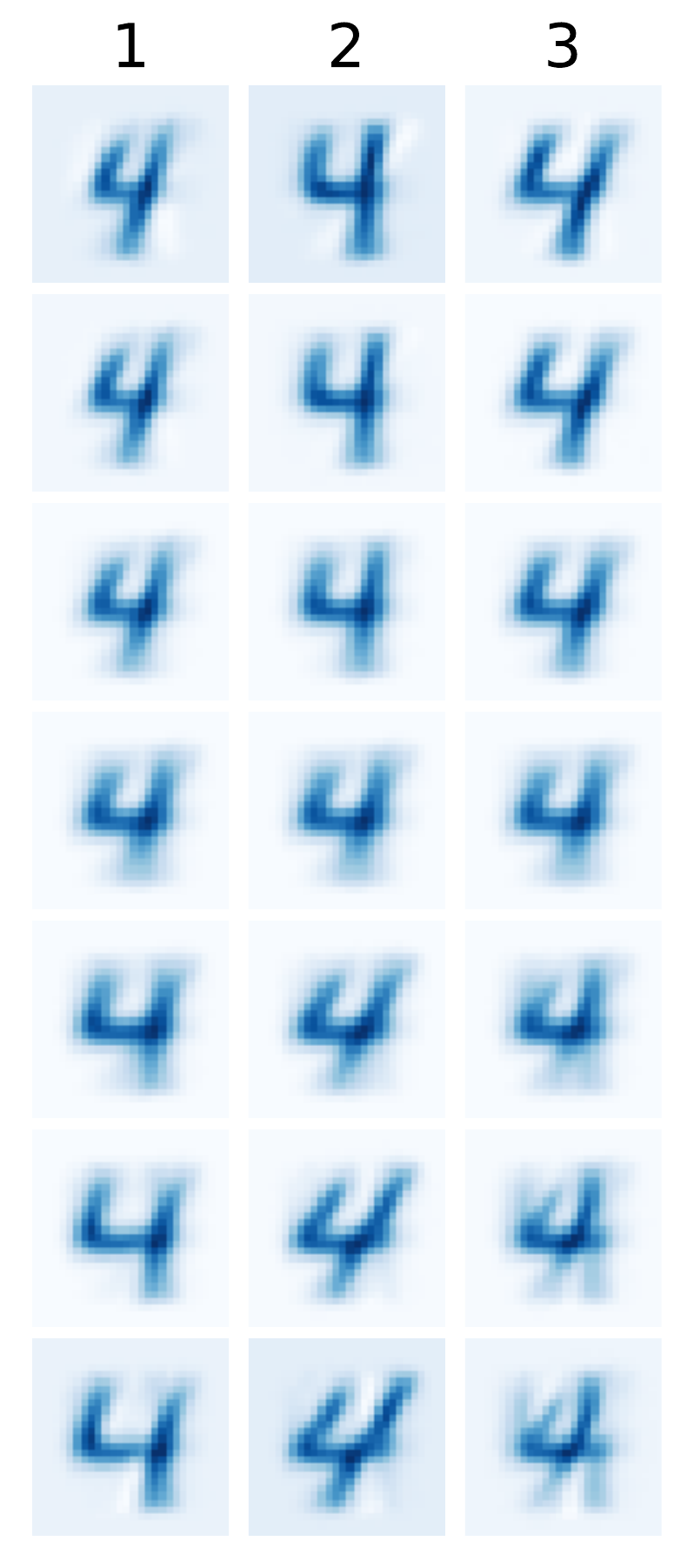}&
  \includegraphics[width=\barywidth\linewidth]{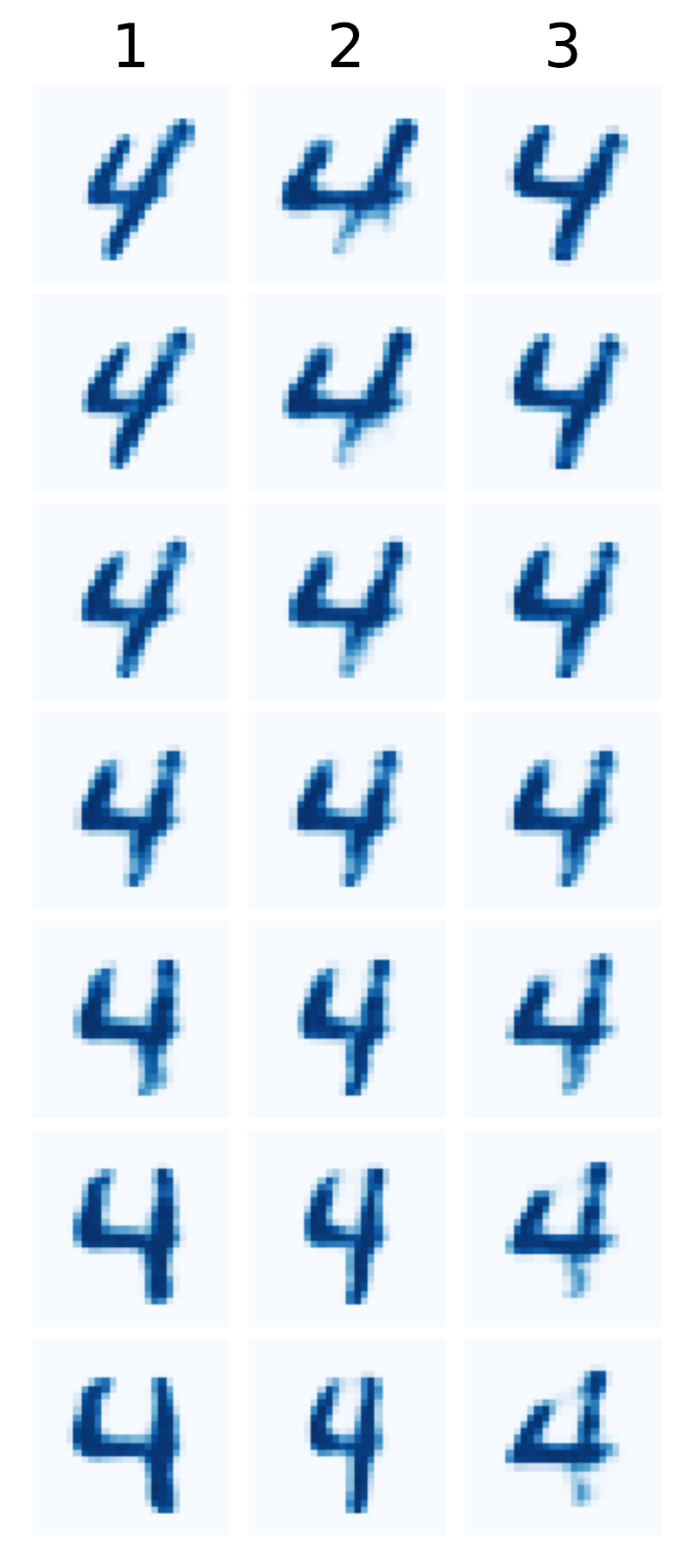}

  \\\hline
\end{tabular}
  \caption{Principal Geodesic Analysis for classes 0,1 and 4 from the MNIST dataset
    for squared Euclidean distance (L2) and Wasserstein Deep Learning (DWE). For
        each class and method we show the variation from the barycenter along one of the first 3
        principal modes of variation.}
  \label{fig:pga}
\end{figure}

\paragraph{Principal Geodesic Analysis} We report in Figure \ref{fig:pga} the
Principal Component Analysis (L2) and Principal Geodesic Analysis (DWE) for 3
classes of the MNIST dataset. We can see that using Wasserstein to encode the
displacement of mass leads to more semantic and nonlinear subspaces such as
rotation/width of the stroke and global sizes of the digits. This is well known
and has been illustrated in~\cite{seguy2015principal}. Nevertheless our method allows for
estimating the principal component even in large scale datasets and our
reconstruction seems to be more detailed compared to~\cite{seguy2015principal}
maybe because our approach can use a very large number of samples
for subspace estimation.

\subsection{Google doodle dataset}

\paragraph{Datasets}
The Google Doodle dataset is a crowd sourced dataset that is freely available from
the web\footnote{\url{https://quickdraw.withgoogle.com/data}} and contains 50 million drawings. 
The data has been collected by asking users to hand draw with a mouse a given
object or animal in less than 20 seconds. This lead to a large number of
examples for each class but also a lot of noise in the sens that people often get
stopped before the end of their drawing .We used the numpy bitmaps format proposed on the quick draw github account. Those are made of the simplified drawings rendered into 28x28 grayscale images.
These images are aligned to the center of the drawing's bounding box. 
In this paper we downloaded the classes Cat, Crab and Faces
and tried to learn a Wasserstein embedding for each of these classes with the
same architecture as used for MNIST. 
In order to create the training dataset we draw randomly 1 million pairs of indexes
from the training samples of each categories and compute the exact Wasserstein distance for quadratic ground
metric using the POT toolbox \cite{flamary2017pot}. Same as for MNIST, 700 000 are used for learning
the neural network, 200 000 are used for validation and 100 000 pars are used for testing purposes.
Each of the three categories( Cat, Crab and Faces) holds respectively 123202, 126930 and 161666 training samples.

\paragraph{Numerical precision and cross dataset comparison} The numerical
performances of the learned models on each of the doodle dataset is reported in
the diagonal of Table \ref{tab:crossperf}.
Those datasets are much more difficult than MNIST because they have not been curated and
contain a very large variance due to numerous unfinished doodles. An interesting comparison
is the cross comparison between datasets where we use the embedding learned on one
dataset to compute the $W^2_2$ on another. The cross performances is given in Table
\ref{tab:crossperf} and shows that while there is definitively a loss in accuracy
of the prediction, this loss is limited between the doodle datasets that all have an
important variety. Performance loss across doodle and MNIST dataset  is
larger because the latter is highly structured and one needs to
have a representative dataset to generalize well which is not the case with MNIST.

\begin{table}[t]\centering
 \begin{tabular}{|c||c|c|c|c|}
 \hline
  Network\/ Data&CAT&CRAB&FACE&MNIST\\
  \hline
  CAT&\textbf{1.195}&\textit{1.718}&2.07&12.132\\
  \hline
  CRAB&\textit{2.621}&\textbf{0.854} &3.158&10.881\\
  \hline
  FACE&\textit{5.025}&5.532&\textbf{3.158} &50.527\\
  \hline
  MNIST&9.118&6.643&\textit{4.68}&\textbf{0.405} \\
  \hline
 \end{tabular}
 \caption{Cross performance between the DWE embedding learned on each datasets.
On each row, we observe the MSE of a given dataset obtained on the deep network learned on the four different datasets (Cat, Crab, Faces and MNIST).}
 \label{tab:crossperf}
\end{table}

\begin{figure}[t]
\centering
\includegraphics[width=.328\linewidth]{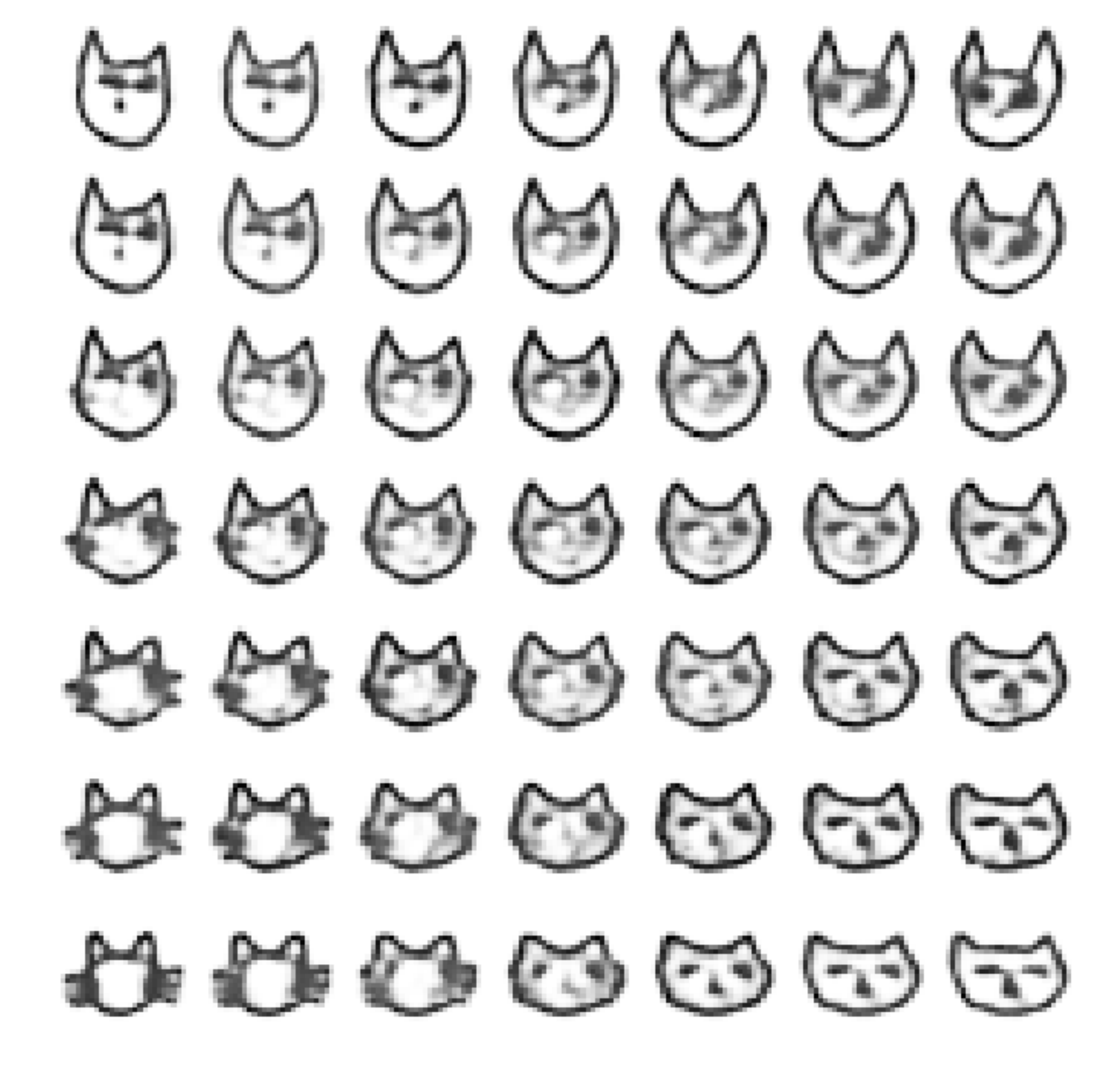}
\includegraphics[width=.328\linewidth]{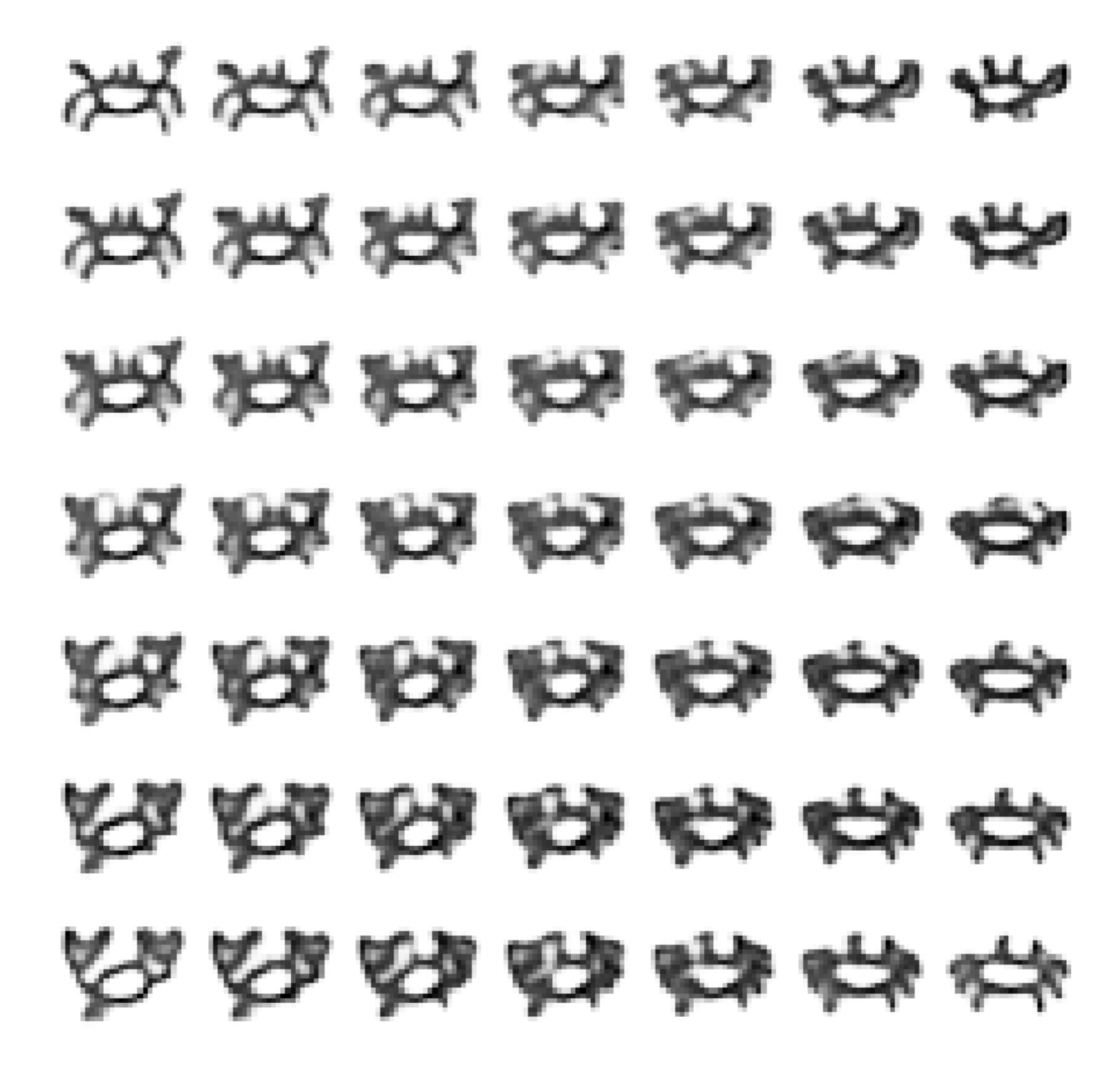}
\includegraphics[width=.328\linewidth]{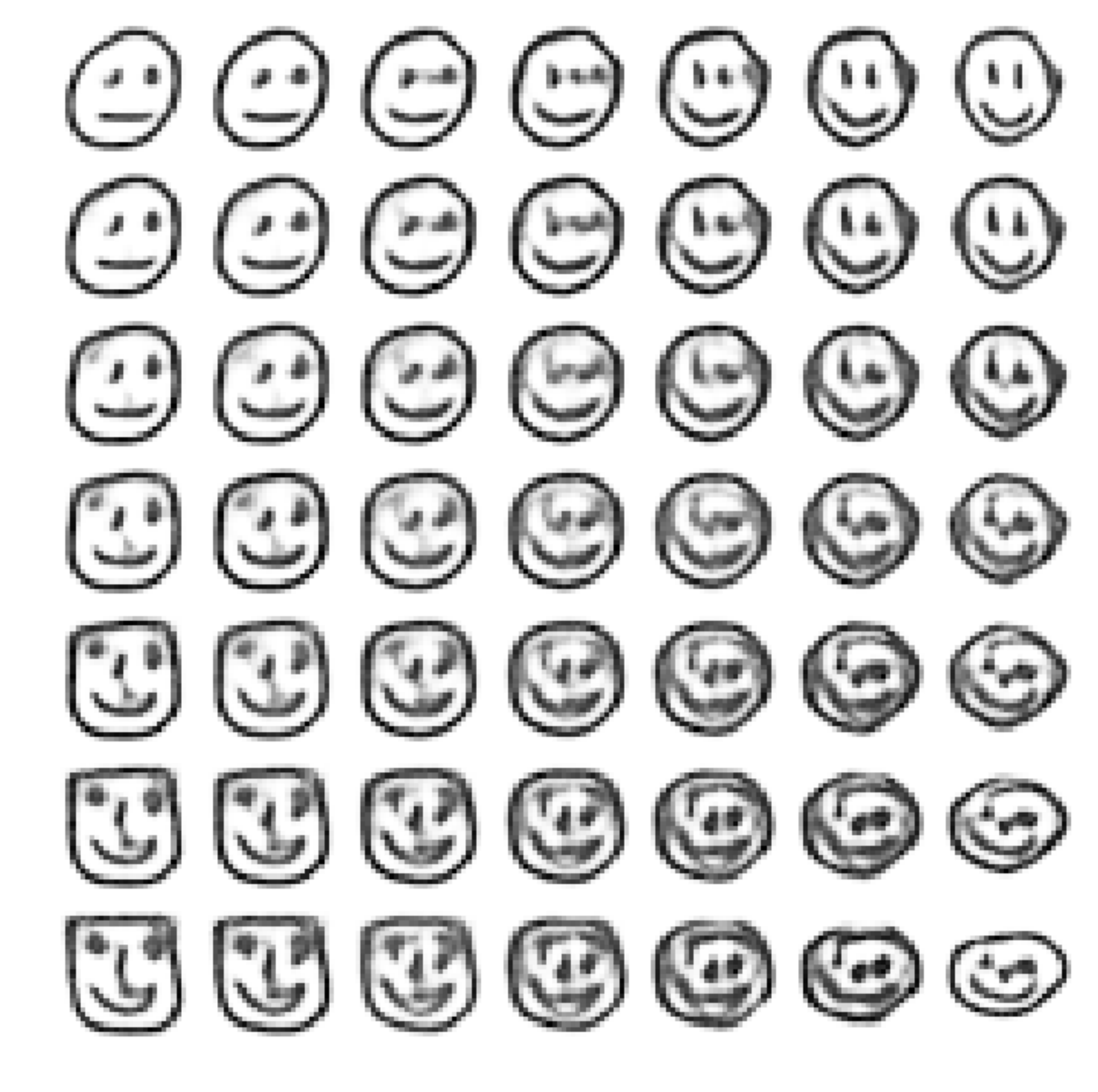}
\caption{Interpolation between four samples of each datasets using DWE.
(left) cat dataset, (center) Crab dataset (right) Face dataset.}
\label{fig:interd2d}
\end{figure}

\paragraph{Wasserstein interpolation} We first compute the Wasserstein
interpolation between  four samples of each datasets in Figure
\ref{fig:interd2d}. Note that these interpolation might not be optimal w.r.t.
the objects but we clearly see a continuous displacement of mass that is
characteristic of optimal transport. This leads to surprising artefacts for
example when the eye of a face fuse with the border while the nose turns into an eye.
Also note that there is no reason for a Wasserstein barycenter to be a realistic
sample.

\begin{figure}[t]
\centering
\includegraphics[width=\linewidth]{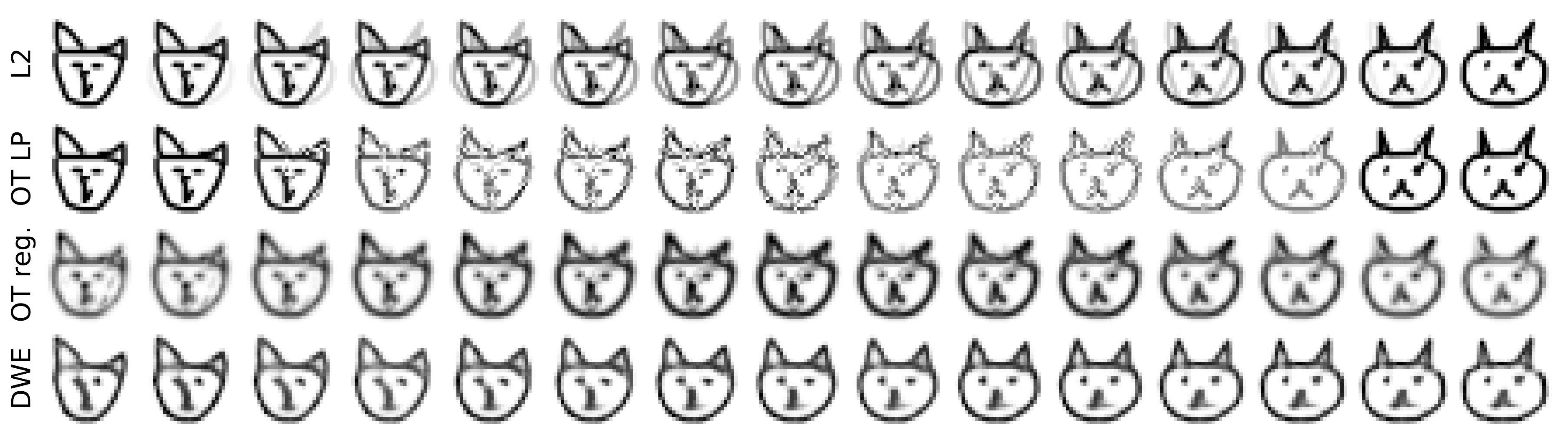}
\caption{Comparison of the interpolation with L2 Euclidean distance (top), LP
Wasserstein interpolation (top middle) regularized Wasserstein Barycenter
(down middle) and DWE (down).}
\label{fig:inter_cat}
\end{figure}

Next we qualitatively evaluate  the subspace learned by DWE by comparing the Wasserstein interpolation
of our approach with the true Wasserstein interpolation estimated by solving
the OT linear program and by using regularized OT with Bregman projections
\cite{benamou2015iterative}. The interpolation results for all those methods
and the Euclidean interpolation are available in Fig.~\ref{fig:inter_cat}.
The LP solver takes a long time (20 sec/interp) and leads to a ``noisy'' interpolation as already explained
in \cite{cuturi2016smoothed}. The regularized Wasserstein barycenter is obtained more
rapidly (4 sec/interp) but is also very smooth at the risk of loosing some
details, despite choosing a small regularization that prevents numerical
problems. Our reconstruction also looses some details due to the Auto-Encoder
error but is very fast and can be done in real time (4 ms/interp).

\section{Conclusion and discussion}

In this work we presented a computational approximation of the Wasserstein distance suitable for large scale data mining tasks. Our method finds an embedding of the 
samples in a space where the Euclidean distance emulates the behavior of the Wasserstein distance. Thanks to this embedding, numerous data analysis tasks can 
be conducted at a very cheap computational price. We forecast that this strategy can help in generalizing the use of Wasserstein distance in numerous applications.

However, while our method is very appealing in practice it still raises a
few questions about the theoretical guarantees and approximation quality. 
First it is difficult to foresee from a given network
architecture if it is sufficiently (or too much) complex for finding a
successful embedding. It can be conjectured that it is dependent on the
complexity of the data at hand and also the locality of the manifold where the
data live in. Second, the theoretical existence results on such Wasserstein
embedding with constant distortion are still lacking. Future works will consider
these questions as well as applications of our approximation strategy on a wider
range of ground loss and data mining tasks. Also, we will study the transferability of one database  to another to diminish the computational burden of computing Wasserstein distances on numerous pairs for the learning process.

 \footnotesize
 \bibliographystyle{abbrv}

\end{document}